\documentclass[letterpaper, 10 pt, conference]{ieeeconf}  % 
\IEEEoverridecommandlockouts                              % This command is only needed if 
\usepackage{amsmath} % assumes amsmath package installed
\usepackage{amssymb}  % assumes amsmath package installed
\usepackage{graphicx}
\usepackage{soul}
\usepackage{xcolor}
\usepackage[acronym]{glossaries}
\usepackage[ruled,vlined,noend]{algorithm2e}
\usepackage{algpseudocode}%
\usepackage{multicol}
\usepackage{cite}
\usepackage{siunitx}
\usepackage{stfloats}
\usepackage{bm}
\usepackage{url}
\usepackage{hyperref}
\usepackage{subcaption}
\usepackage[compact]{titlesec}         % you need this package
\titlespacing{\section}{0pt}{0pt}{0pt} % this reduces space between (sub)sections to 0pt, for example
\AtBeginDocument{%                     % this will reduce spaces between parts (above and below) of texts within a (sub)section to 0pt, for example - like between an 'eqnarray' and text
  \setlength\abovedisplayskip{0pt}
  \setlength\belowdisplayskip{0pt}}

\title{\LARGE
SPADE: Towards \textbf{S}calable \textbf{P}ath Planning Architecture on \textbf{A}ctionable Multi-\textbf{D}omain 3D Scen\textbf{E} Graphs.
}

\author{Vignesh Kottayam Viswanathan, Akash Patel, Mario A.V. Saucedo, Sumeet Gajanan Satpute,\\ Christoforos Kanellakis and  George Nikolakopoulos% <-this % stops a space
\thanks{The authors are with Robotics and AI, Luleå University of Technology 97187, Luleå,Sweden
        {\tt\small \{vigkot, akapat, marval, sumsat, chrkan and geonik\}@ltu.se}}%
\thanks{This work has been partially funded by the European Union's Horizon Europe Research and Innovation Programme, under the Grant Agreement No.101138451 PERSEPHONE.}
}

\begin{document}

\maketitle
\thispagestyle{empty}
\pagestyle{empty}

%%%%%%%%%%%%%%%%%%%%%%%%%%%%%%%%%%%%%%%%%%%%%%%%%%%%%%%

\begin{abstract}
In this work, we introduce SPADE, a path planning framework designed for autonomous navigation in dynamic environments using 3D scene graphs. SPADE combines hierarchical path planning with local geometric awareness to enable collision-free movement in dynamic scenes. The framework bifurcates the planning problem into two: (a) solving the sparse abstract global layer plan and (b) iterative path refinement across denser lower local layers in step with local geometric scene navigation. To ensure efficient extraction of a feasible route in a dense multi-task domain scene graphs, the framework enforces informed sampling of traversable edges prior to path-planning. This removes extraneous information not relevant to path-planning and reduces the overall planning complexity over a graph. Existing approaches address the problem of path planning over scene graphs by decoupling hierarchical and geometric path evaluation processes. Specifically, this results in an inefficient replanning over the entire scene graph when encountering path obstructions blocking the original route. In contrast, SPADE prioritizes local layer planning coupled with local geometric scene navigation, enabling navigation through dynamic scenes while maintaining efficiency in computing a traversable route. We validate SPADE through extensive simulation experiments and real-world deployment on a quadrupedal robot, demonstrating its efficacy in handling complex and dynamic scenarios.

\end{abstract}

%%%%%%%%%%%%%%%%%%%%%%%%%%%%%%%%%%%%%%%%%%%%%%%%%%%%%%%

\section{Introduction}
% The recent emergence of 3D scene graphs as an intuitive, versatile, and tractable representation of environments has driven groundbreaking advancements in autonomous robotics. Applications span across, but not limited to, autonomous driving~\cite{yu2021scene,greve2024collaborative}, exploration~\cite{liu2022explore,zhu2023excalibur}, scene navigation~\cite{ravichandran2022hierarchical,singh2023scene,werby2024hierarchical,fadhil2024seek}, and planning~\cite{agia2022taskography,rana2023sayplan,gu2024conceptgraphs,dai2024optimal}. Overall, scene graphs provide a information-rich and actionable world model that can be effectively manipulated to achieve diverse mission objectives.

In this work, we focus on utilizing an actionable hierarchical scene representation to tackle the task of path planning. Planning the shortest, collision-free path from point A to point B is a fundamental problem in robotics, essential for real-world applications. Classical approaches frame this problem within a geometric model, where traversable space can be represented as a discretized grid or as a graph. This is then processed by a corresponding path planner to generate a feasible route~\cite{karlsson2023d+,patel2024stage,lindqvist2024tree}. However, when managing long-term, persistent representation of large-scale environments, in contrast to directly planning on volumetric representation, initial planning over abstractions can reduce the overall computational overhead required for subsequent geometric planners~\cite{larsson2020q,rosinol2021kimera,ray2024task}.

\begin{figure}[htbp]
    \centering
    \includegraphics[width=\linewidth]{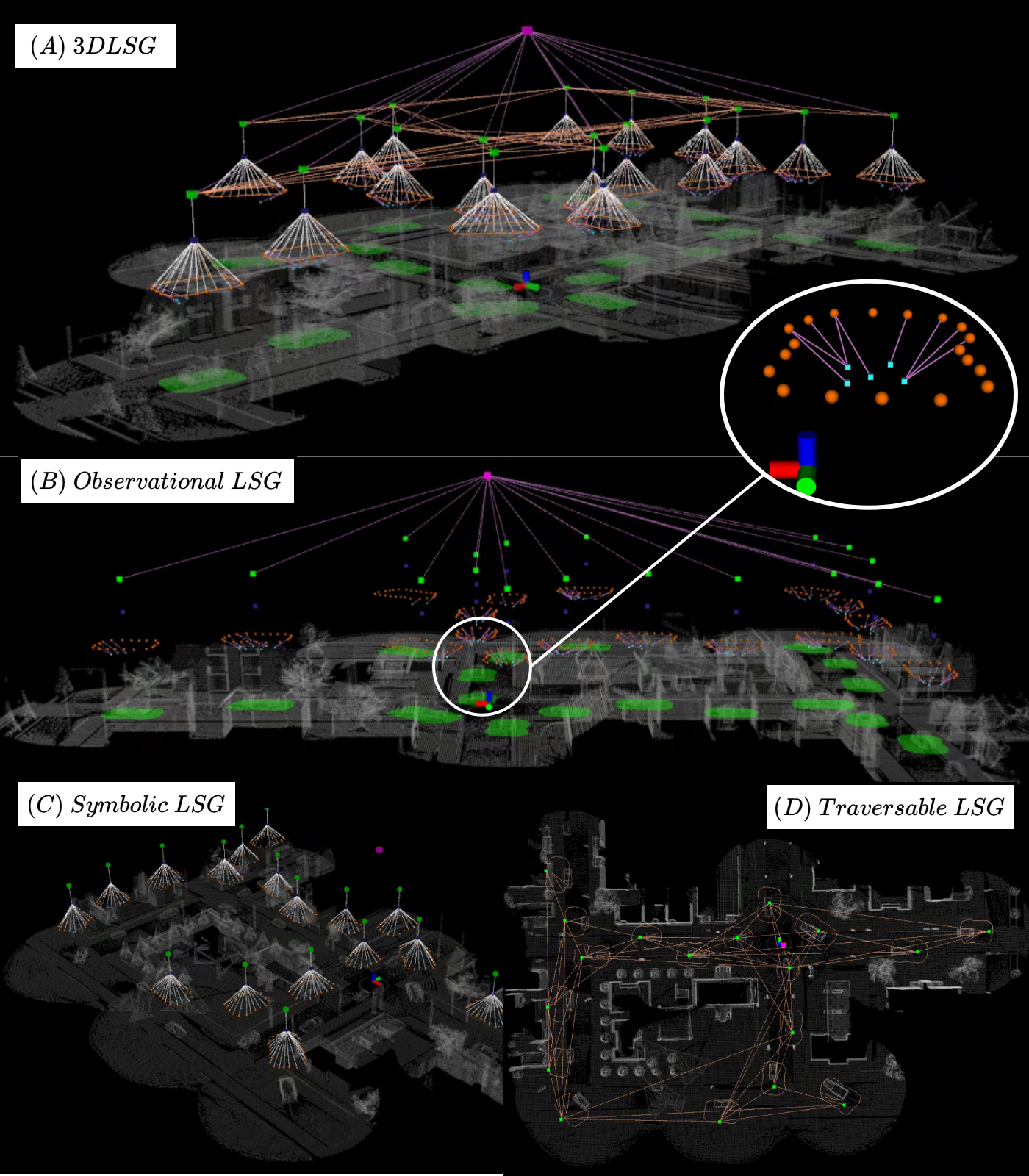}
    \caption{A graphical abstract portraying a multi-domain 3D scene graph that could be used to address diverse tasks based on its relevant intrinsic composition.}
    \label{fig:SPADE_lsg}
\end{figure}

Evaluating a traversable route over an actionable 3D scene graph is generally approached in a hierarchical manner, i.e. parsing through corresponding graph representation of abstract layers in a cascading \textit{top-down} manner. The global semantic-metric path is then shared with a conventional volumetric-based planner to generate a final collision-free path for the robot to track. As scene graphs continuously evolve to support multiple task domain operations such as path planning~\cite{rosinol2021kimera,kremer2023s,ray2024task}, object search~\cite{ge2024commonsense}, reasoning~\cite{Saucedo2024BeliefSG,kim20193}, their intrinsic composition grows increasingly complex. Intra-layer edges no longer represent only the existence of traversable paths but also encode semantic relationships and  affordance mappings between nodes. This results in scene graphs becoming increasingly \textit{dense} with extraneous information, in terms of computing a path, as the robot explores an environment. Figure.~\ref{fig:SPADE_lsg} demonstrates an example of multi-domain 3D scene graph, a modified Layered Semantic Graphs (LSG)~\cite{viswanathan2024xflie} used in this work to address the task of path-planning. Figure.~\ref{fig:SPADE_lsg}(B-D) highlights the various types of sub-graphs that could be potentially extracted from Figure.~\ref{fig:SPADE_lsg}(A) based on the relevant task.

Thus, addressing navigational queries over 3D scene graphs presents two main challenges. The \textbf{first challenge} lies in the performance of global planning at scale, which is affected by both the size of the scene graph and the inherent traversability uncertainties encountered when planning a global route in real-world conditions. Specifically, graphs within lower abstraction layers become more densely populated as the robot explores newer regions. This increases the graph size that needs to be processed by the planner as it parses through the layers. Moreover, in large-scale environments, real-world conditions may alter the initially registered traversability between nodes. Such changes can invalidate the original navigation plan when the robot detects and updates node connections to reflect path obstructions upon reaching them. Since a change in traversability condition is likely for nodes beyond the robot's perception range, overlooking this possibility during the initial global planning stage may necessitate replanning through the necessary abstraction layers, resulting in increased computational overhead. The \textbf{second challenge} is the presence of extraneous intra-layer edge relations in a multi-domain scene graph. These relations result in \textit{uninformed} evaluations, where standard graph-based path planners naively process all non-domain specific relations. This increases the search space and ultimately impacts computational efficiency.

To address these challenges, we develop and demonstrate a framework that focuses on fast, recursive planning on local graph representation of lower abstraction layers while ensuring global route awareness by restricting planning on the global graph representation to the highest abstraction layer. Through this bifurcation, we argue that the planning performance would benefit positively by actively limiting the exposure of the planner to size of the graph that it needs to plan over. Our framework, additionally, implements a prior step of \textit{informed} evaluation of intra-layer connections to efficiently perform domain-specific actions, in this case: path planning, over large-scale 3D scene graphs. Thus, the proposed developments, together comprise \textbf{SPADE}, a full-fledged path planning framework leveraging hierarchical 3D scene graphs to address semantic and onboard autonomy navigational queries.

\section{Related Works}

In view of current state-of-art articles, limited works exist that investigate the technical implementation of path planning on 3D scene
graphs.~\cite{fernandez1998hierarchical} investigated path planning over hierarchical graphs to navigate through an environment. In their work, the authors adopt a recursive planning strategy which cascades over the depth of the established hierarchy to refine the resolution of the planned path.~\cite{rana2023sayplan} integrates classical path planning technique with LLM-derived high-level guidance to navigate an indoor
scene graph. In their work, the authors utilize a graph-based planner to handle semantic route planning tasks within the generated high-level plan.
~\cite{rosinol2021kimera,kremer2023s} demonstrates the advantage of hierarchical path planning to address semantic navigational queries. In~\cite{rosinol2021kimera}, the authors present a decoupled and cascading pipeline where an initial high-level global route is extracted over the abstraction layers. Similar to~\cite{fernandez1998hierarchical}, the high-level route definition is refined as the planner iterates over deeper abstraction layers. The final route is then subsequently shared with a geometric planner that operates within local perception bounds of the robot in order to generate a smooth and collision-free path.~\cite{ray2024task} investigates the problem of task and motion planning over hierarchical representation such as 3D scene graphs. In their work, the authors extract planning abstraction required to solve a given task over the scene graph and subsequently generate a semantic-metric path to reach the query node. Finally, this is used to generate a kinematically feasible navigational solution by a sampling based planner to move the robot through the scene.

\subsection{Contributions}

%Based on the related state-of-art works in the direction of path planning on 3D scene graphs, we present the technical contributions comprising the proposed SPADE framework in this section.  

The first contribution of SPADE stems from a novel hierarchical path planning framework designed to efficiently navigate large-scale scene graphs. Given a navigational query, SPADE computes an initial global path across the highest abstraction layer and segments the path. Each high-level path segment is sequentially shared with the planner nested in the lower abstraction layers to generate a refined plan, enabling navigation through the layers. By coupling the high-level local layer path with a volumetric local path planner, we restrict the planner to operate within the bounds of the local representation, in the sense of both abstract and geometric scene representation. Through this approach, we actively address the redundant problem of generating a global semantic-metric route when considering presence of traversability uncertainty in dynamic large-scale environments. Simultaneously, this approach also limits the overall graph size exposed to the planner for a single planning instance. 
%  Existing works approach this problem by first determining the global route over the scene graph and subsequently refining the solution in a geometric-representation grounded planner. When encountering obstructions en-route, a global replanning is initiated, either across the entire scene graph or to address a specific segment traversal. In both cases, the planning instance is evaluated a minimum of two times even under the consideration of traversability uncertainty in dynamic environments. In contrast to that, we propose a synchronous planning solution which aims to prioritize finding a navigational traversing the current local layer representation. 

The second contribution of SPADE is derived from a novel informed sampling method which enables efficient parsing of domain-specific edges. This development address the computational bottleneck during parsing multi-domain scene graph representation to address a navigational query. To the best of authors' knowledge, no prior works exist that consider the presence of diverse nature of edge-relations in a scene graph, especially while addressing the problem of path planning. The proposed sampling method is evaluated a step prior to evaluation of the graph by a graph-based planner. This is achieved by considering the context of the query, i.e. navigation, which is then utilized to extract a temporary topological sub-graph of the current layer representation containing the required edge relations. The extracted sub-graph with relevant edge attributes is then shared with the graph-based planner to further extract a traversable route.

Together, these contributions position SPADE as a scalable solution to address path planning queries by leveraging actionable 3D scene graphs.

\section{SPADE Planner Architecture}

\begin{figure*}[ht]
    \centering
    \includegraphics[width=\linewidth]{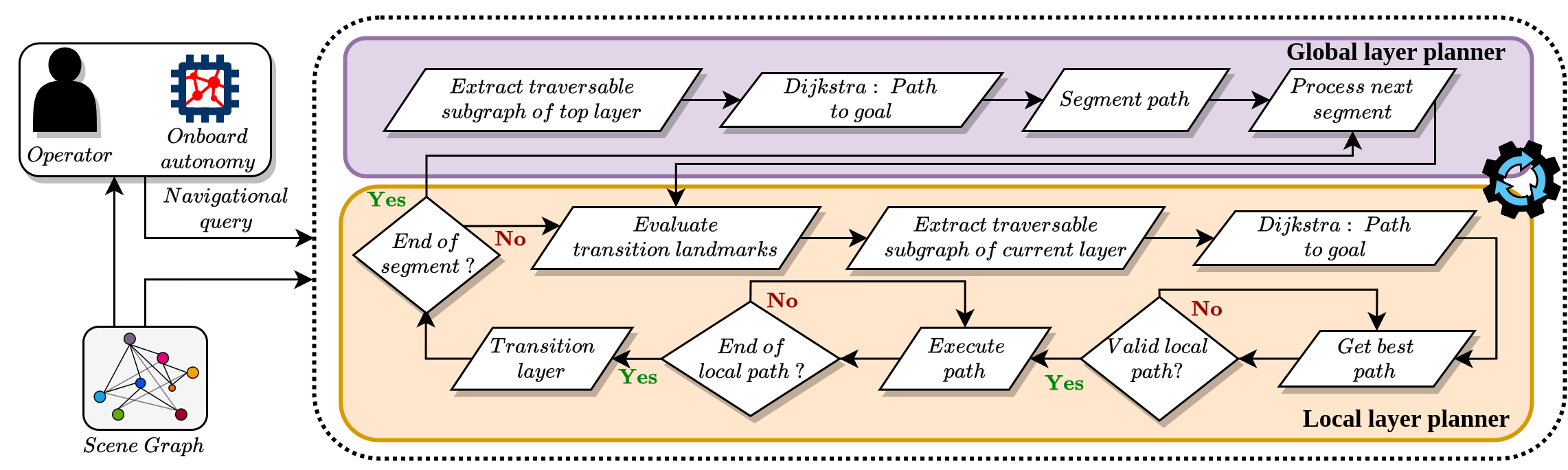}
    \caption{A high-level overview of the proposed SPADE planner architecture to address path planning over 3D scene graphs.}
    \label{fig:SPADE_architecture}
\end{figure*}

Figure.~\ref{fig:SPADE_architecture} presents a high-level overview of the proposed path planning architecture over 3D scene graphs. As an input to the architecture, SPADE considers the navigational task shared either by an operator as a semantic query, i.e. $``$ \texttt{Inspect} \textit{open-window} in \textit{Level-2} of \textit{Building-1}$"$ or directly by the onboard autonomy, i.e. $``$\texttt{GoTo} \textit{Target-car-3}$"$, during mission planning. We denote $\textbf{X}_{odom} \in \mathbb{R}^4 | \textbf{X}_{odom} = [x,y,z,\psi]$ as the localization estimate of the robot provided by the onboard Simultaneous Localization and Mapping (SLAM) module. Furthermore, we define $\mathbb{G} = (V,E)$ as the scene graph entity composed of nodes $v \in V$ and edges $e \in E$ across the abstraction layers $l \in L$. We derive the scene graph representation required for evaluating the proposed architecture from~\cite{viswanathan2024xflie}. In~\cite{viswanathan2024xflie}, the authors define and construct a nested actionable 3D Layered Semantic Graphs (3DLSG) during inspection. The readers are directed to~\cite{viswanathan2024xflie} for a detailed understanding of 3DLSG and its construction. Briefly, the 3DLSG in consideration is composed of four main layers, mainly:

\begin{itemize}
    \item \textbf{Target} layer: The highest abstraction layer maintaining a graph $\mathbb{G}_T$ populated with nodes ($v_T \in V$) capturing observed inspection targets in the scene. 
    \item \textbf{Level} layer: A lower abstraction layer maintaining a graph $\mathbb{G}_L$ populated with nodes ($v_L  \in V$) representing the inspection levels maintained during the coverage of registered target node. This layer is local and nested within every registered $v_T$.
    \item \textbf{Pose} layer: An abstraction layer maintaining a graph $\mathbb{G}_P$ populated with the nodes ($v_P  \in V$) capturing the maintained inspection view-pose configuration for every level of inspection of an observed target. This layer is local and nested within every registered $v_L$.
    \item \textbf{Feature} layer: An abstraction layer maintaining a graph $\mathbb{G}_F$ populated with the nodes ($v_F  \in V$) representing the observed semantic features from each view-pose configuration. This layer is local and nested within every registered $v_P$.
\end{itemize}

Each node comprises of an attribute set $n_v \in N_V$ and each edge comprises of an attribute $n_e \in N_E$. For the purpose of this work, $n_v$ typically comprises of the 3D pose of the node ($\bm{X}_v \in \mathbb{R}^3 | \bm{X}_{v_l} = [x,y,z]$) and its layer identification ($l$). However, based on the parent layer of the node, the node attributes can include the corresponding lower layer graph representation ($\mathbb{G}_{l}$). Furthermore, $v_T$ nodes can additionally encode the inspection status ($I \in \mathbb{R}^+$), the determined level transition node ($v_L$) to enable navigation planning between $v_T$ nodes. We additionally account for the presence of multiple edge relations between nodes which can be encoded with individual, purpose-specific information. Specifically, we consider three main types of edge relations possible between nodes mainly $n^s_e$ which encodes symbolic relation between parent-child nodes, $n^t_e$ which encodes traversability information such as the weighted traversable cost ($w_t \in \mathbb{R}^+$) and the semantic status of traversability ($\mathcal{T}$) and finally, $n^o_e$ which encodes observational relation such as the confidence score ($\gamma \in \mathbb{R}^+$) of an object detection approach (e.g. YOLO\cite{yolo11_ultralytics}) as well as the distance to the observed object ($w_o \in \mathbb{R}^+$). Thus, based on the available edge types we further define $\kappa \in \texttt{TaskMode}~|~\texttt{TaskMode} = [\textit{path finding},\textit{anomaly detection}, \textit{scene understanding}]$ as the planning actions to be considered possible over the scene graph. We denote $\{v^{trm}_l\} \in V$ as the relevant set of terminal nodes parsed from the navigational query, representing the goal nodes across the abstraction layers. Additionally, we define $\{v^{curr}_l\} \in V$ as the corresponding subset of nodes representing the current location of the robot across the abstraction layers. 

To ensure collision-free navigation, we employ a modified version of STAGE~\cite{patel2024stage} that operates on voxelized environment representation ($\mathbb{M}$). STAGE is a graph-based planner designed to enable autonomous exploration. We modify the core capability of STAGE to additionally support query-based path execution as required in this work. Based on current 3D pointcloud $\{P\} \in \mathbb{R}^3$ and $\bm{X}_{odom}$, STAGE leverages local 3D voxelized environment representation ($\mathbb{M}_L$) built through \textit{voxblox}~\cite{oleynikova2017voxblox} to processes, on-demand, the initial high-level route into a refined collision-safe path which can be tracked by the robot. For more information on the functionality of STAGE, the readers are directed to~\cite{patel2024stage}.

\subsection{Domain-aware Graph Subsampling}\label{sec:dags}

\begin{figure}[htbp]
    \centering
    \includegraphics[width=\linewidth]{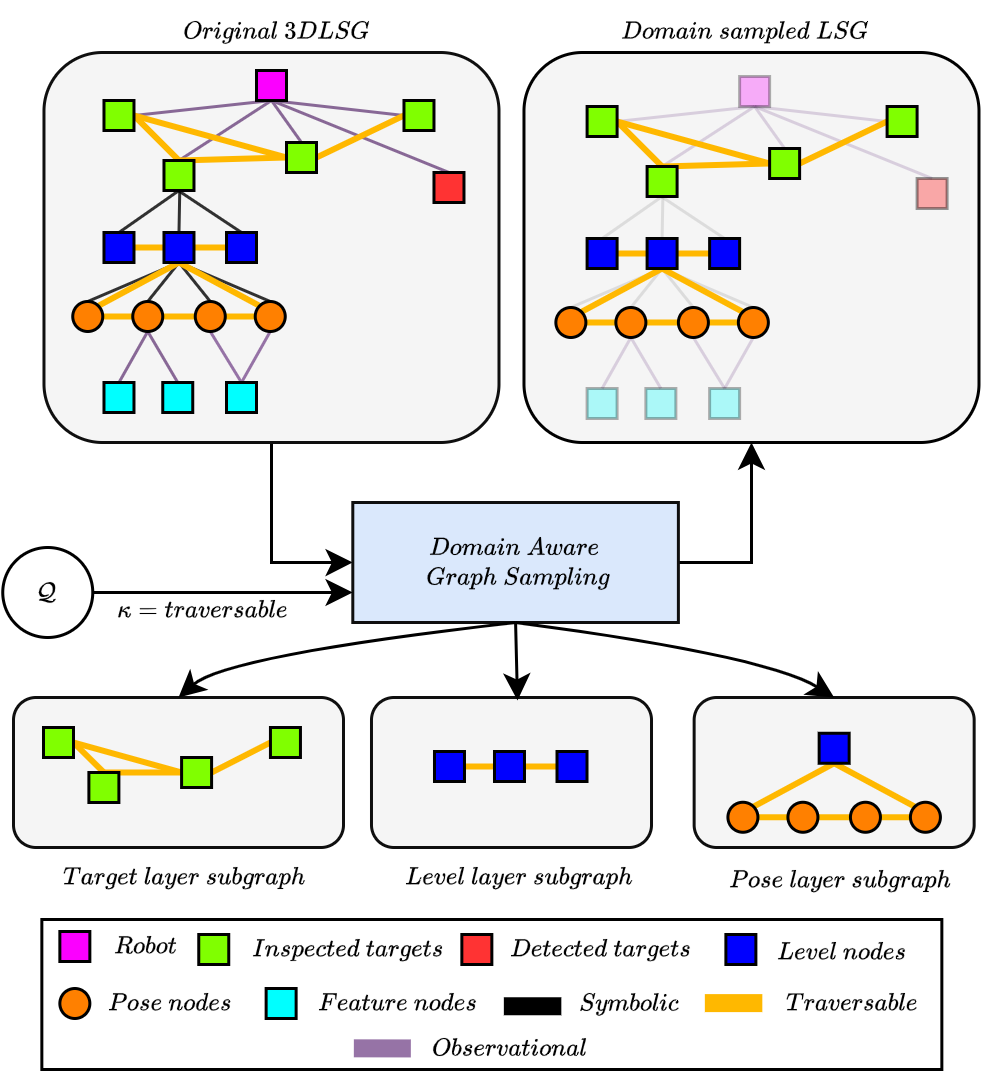}
    \caption{A visual overview of the domain aware graph sampling approach presented to subsample the multi-domain 3D scene graph.}
    \label{fig:dags}
\end{figure}

In this article, we focus on the task of $\textit{path-finding}$ as the desired planning action to remain within the scope of this work. Figure~\ref{fig:dags} visually highlights the graph subsampling process in response to the task at hand. Based on $\kappa$, the relevant edge ($\{e^{\mathbb{S}}_l
\} \subseteq E_l \in \mathbb{G}_l$) and the corresponding edge-attribute ($\{n^t_e\}$) subsets are extracted from the initial $\mathbb{G}_l$. This is achieved by iterating over and filtering the node and edge composition of $\mathbb{G}_l$. Once $\{e^{\mathcal{S}}_l\}, \{n^t_e\}$ is made available, the relevant actionable sub-graph representation $\mathbb{G}^{\mathcal{S}}_l \subseteq \mathbb{G}_l$ is built. Subsequently, $\mathbb{G}^{\mathcal{S}}_l$ is shared with a graph-based path planner such as Dijkstra's~\cite{dijkstra1959note} for further planning. This process is then called and executed for each and every graph representation of an abstraction layer that needs to be traversed in order to reach the goal. 

% \begin{algorithm}
%     \caption{Domain-aware Graph Subsampling}\label{alg:DAGS}

%     \SetKwComment{Comment}{/* }{ */}
%     \KwIn{$\mathbb{G}_{l}$, \texttt{TaskMode}}

%     $\mathbb{G}^{\mathcal{S}}_l \gets \textbf{InitializeSubGraph}()$

%     $\kappa \gets \textbf{ParseTaskAttribute}(\texttt{TaskMode})$
    
%     $\{e^\mathcal{S}_l\}, \{n^{\kappa}_e\} \gets \textbf{ExtractValidConnections} (\mathbb{G}_l,\kappa)$

%     $\mathbb{G}^\mathcal{S}_l \gets \textbf{BuildSubGraph}(\{e^\mathcal{S}_l\},\{n^{\kappa}_e\}) $

%     % $\textbf{RunHierarchicalPathPlanner}(\mathbb{G}^{\mathcal{S}}_l,\mathcal{Q})$
% \end{algorithm}

\subsection{Semantic-Geometric Path Planning}\label{sec:sgpp}
\begin{algorithm}
    \caption{Semantic-Geometric Path Planner}\label{alg:SCOUT}

    \SetKwComment{Comment}{/* }{ */}
    \SetKwFunction{isCollisionFree}{isCollisionFree}

    \KwIn{$\mathbb{G}_{T}$,\texttt{TaskMode},$\{v^{trm}_l\}$}

    $v^{curr}_T,v^{curr}_L,v^{curr}_P \gets \textbf{LocalizeRobotInGraph}(\mathbb{G}_{T})$  
    
    $\mathbb{G}^{\mathcal{S}}_T \gets \textbf{ExtractDomainSubgraph}(\mathbb{G}_{T},\texttt{TaskMode})$
    
    $\{\Pi^G\} \gets \textbf{EvalGlobalLayerPath}(\mathbb{G}^{\mathcal{S}}_{T},v^{curr}_T,v^{trm}_T)$

    $\Pi^{G*} \gets \textbf{GetBestPath}(\{\Pi^G\})$

    $\pi^G_i \gets \textbf{SegmentPath}(\Pi^{G*})$

    $\Hat{\pi}^L \gets \textbf{EvalLocalLayerPath}(\pi^{G}_i,v^{curr}_T,v^{curr}_L,v^{curr}_P)$
    $\textbf{EvalSegmentTraversability}(\mathbb{M}_L,\pi^L)$
    
    \If{$\textbf{not}~\isCollisionFree(\pi^L)$}{
    $\lambda^k \gets \textbf{AltLocalGeometricPathSegment}(\pi^L)$
    
        \If{$\lambda^k \gets \emptyset$}{
        $\mathbb{G}^S_l \gets \textbf{UpdateLocalLayerGraph}(\mathbb{G}^S_l)$
        $\Hat{\pi}^L \gets \textbf{ReplanLocalLayerPath}(\mathbb{G}^S_l)$
        
            \If{$\pi^L \gets \emptyset$}{
            $G^S_T \gets \textbf{UpdateGlobalLayerGraph}(\mathbb{G}^S_T)$
            $\{\Hat{\Pi}^{G}\} \gets \textbf{ReplanGlobalLayerPath}(\mathbb{G}^S_T)$
            $\Hat{\Pi}^{G*} \gets \textbf{GetBestPath}(\{\Hat{\Pi}^G\})$
            $\Hat{\pi}^G_i \gets \textbf{SegmentPath}(\Hat{\Pi}^{G*})$
            
            $\pi^L \gets \textbf{EvalLocalLayerPath}(\Hat{\pi}^{G}_i,v^{curr}_T,v^{curr}_L,v^{curr}_P)$
            }
        }
    }
    
    $\textbf{EvalSegmentTraversability}(\mathbb{M}_L,\pi^L)$
    $\lambda^k \gets \textbf{EvalGeometricPathSegment}(\pi^L)$ 
    $\textit{nMPC} \gets \textbf{ExecuteGeometricPath}(\lambda^k)$

\end{algorithm}

Subsequent to the extraction of domain-specific actionable sub-graph, presented in Sec.~\ref{sec:dags}, the hierarchical path planning operation is executed to move towards $\{v^{trm}_l\}$. The proposed hierarchical path planning framework is composed of two main stages: (a) the initial stage where a route across the global graph of the highest abstraction layer is evaluated and (b) the second stage where the planner extracts a feasible route over the local graph representation of the lower abstraction layers to traverse through to each intermediate higher abstraction node within the planned global route. Algorithm.~\ref{alg:SCOUT} presents the implementation of the semantic-geometric planning for the evaluation of the current global route segment. The planned path is then segmented into its individual pose references and shared with the onboard nonlinear Model Predictive Controller (nMPC) to track. We utilize the implementation of nMPC as presented here~\cite{karlsson2022ensuring}. This process is recursively executed until the robot arrives at the queried terminal nodes.

\textbf{Global Layer Planning:} We denote $\{\Pi^G\}$ as the computed set of available routes by Dijkstra's algorithm to reach $v^{trm}_{T}$ from $v^{curr}_{T}$ through the global graph representation of the highest abstraction layer. For further planning, the shortest route $\Pi^{G*} \in \{\Pi^{G}\}$ is considered. Subsequently, as depicted in algorithm.~\ref{alg:SCOUT}, the corresponding path segments $\pi^{G}_i \in \Pi^{G*} | \pi^G_i = \{v_{T,i},v_{T,i+1}\}$ is iteratively processed at the second planning stage to extract a feasible route through the local abstraction layers of the scene graph. For each $\pi^{G}_i$, the planner begins by evaluating the available layer transition landmark nodes ($v^{tran}_l \in V$) in the \textit{Level} and \textit{Pose} layers. These transition nodes serve as connectors between adjacent local layers, allowing the planner to move beyond the current layer and access the local sub-graph representation of the next layer for continued planning. Since 3DLSG is a nested scene graph structure—where each lower-layer representation is encapsulated within a node of the parent layer—neighbouring graph representations are not immediately accessible for direct, single-shot planning. Instead, the planner must first identify and reach an appropriate layer transition node. Only upon reaching this node can it shift to the next layer, instantiate the relevant local graph representation, and proceed with planning toward the next target node $v_T$ in the sequence $\pi^{G}_i$.

\begin{figure}[htbp]
    \centering
    \includegraphics[width=\linewidth]{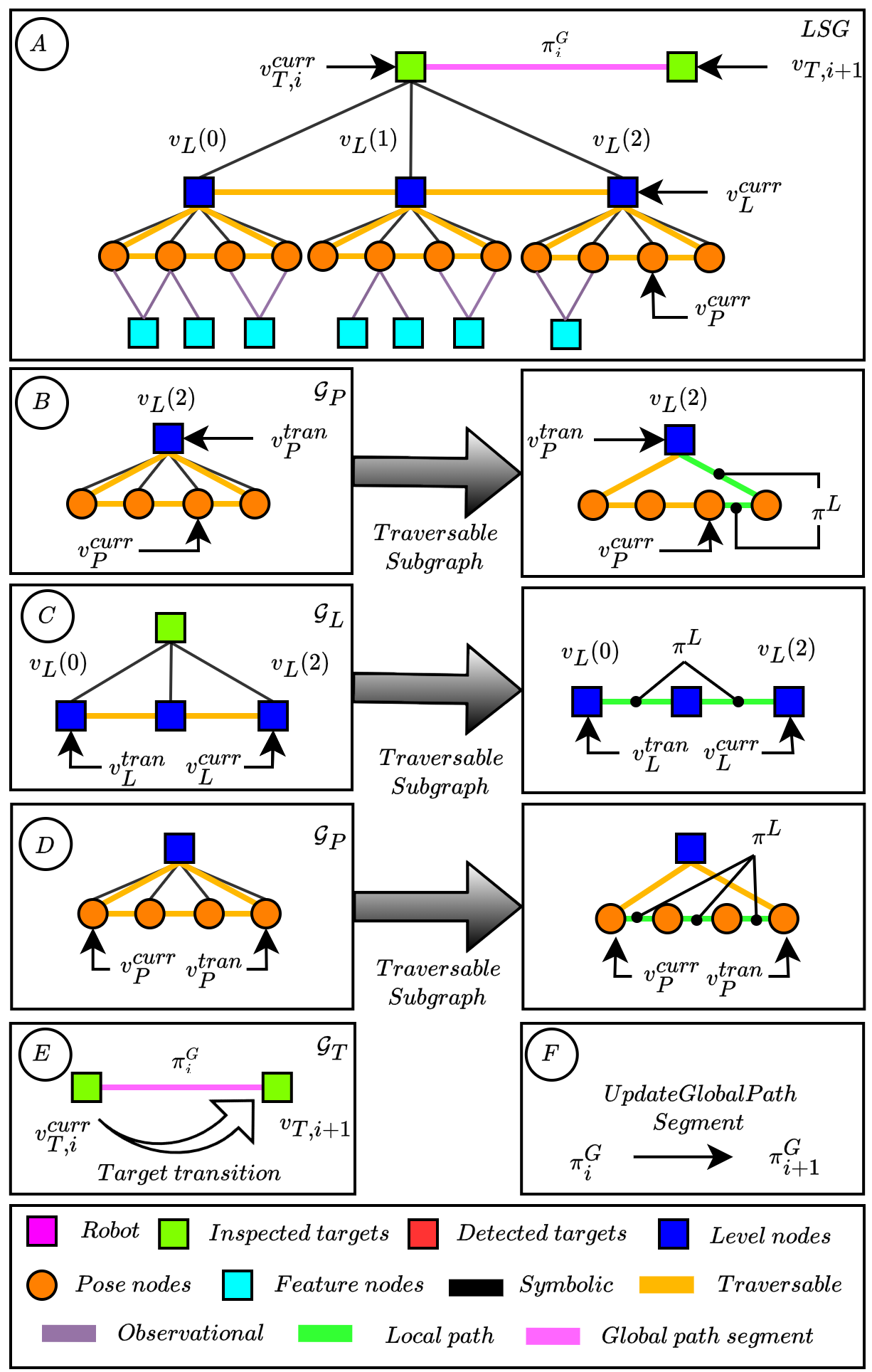}
    \caption{An overview of the local layer planning process during evaluation of the current global path segment over the nested 3D LSG.}
    \label{fig:local_layer_planner}
\end{figure}

\textbf{Local Layer Planning} Once $\{v^{tran}_l\}$ is evaluated and the robot is localized within 3DLSG, the local layer path ($\pi^{L}_l \in \mathbb{G}_l| \pi^{L}_l = \{v_l\} $) is planned through the scene graph. Unlike existing works~\cite{rosinol2021kimera}, which follow a coarse-to-fine planning approach by iterating through global abstraction layers from the highest to the lowest to generate a traversable path, we adopt a planning strategy that is not constrained by a preset sequence. Instead, at each planning instant, we dynamically determine the appropriate layer representation to traverse based on the evaluated transition nodes ($v^{tran}_L, v^{tran}_P$) and the current nodes ($v^{curr}_L, v^{curr}_P$) where the robot is localized. Figure.~\ref{fig:local_layer_planner} presents a visual breakdown of the local layer planning pipeline presented earlier in Fig.~\ref{fig:SPADE_architecture}. Fig.\ref{fig:local_layer_planner}(\textbf{A}) captures the robot begin localized within the layers of 3DLSG after receiving $\pi^G_i$ for traversal.

% For example, if $v^{tran}_L \neq v^{curr}_L$, then the planner first computes $\pi^\mathcal{L}_P$ through the expanded $\mathbb{G}_P$ representation of $v^{curr}_L$ to reach $v^{tran}_P$. At $v^{tran}_P$, the planner transitions from the \textit{Pose} layer into the \textit{Level} layer abstraction, updating the current tangible graph representation to $\mathbb{G}_L$. Subsequently, the planner computes $\pi^\mathcal{L}_L$ to reach $v^{tran}_L$ from $v^{curr}_L$. Furthermore, if there is a queried $v^{trm}_P$ at $v^{tran}_L$, the planner subsequently transitions from \textit{Level} layer into the \textit{Pose} layer representation, updating the current tangible graph representation to $\mathbb{G}_P$ for further planning.  

For example, if $v^{tran}_L \neq v^{curr}_L$, the planner first computes $\pi^{L}_P$ within the expanded $\mathbb{G}_P$ representation of $v^{curr}_L$ to reach $v^{tran}_P$ (refer Fig.~\ref{fig:local_layer_planner}(\textbf{B})). Upon arriving at $v^{tran}_P$, the planner transitions from the \textit{Pose} layer to the \textit{Level} layer, updating the tangible graph representation to $\mathbb{G}_L$. It then computes $\pi^{L}_L$ to navigate from $v^{curr}_L$ to $v^{tran}_L$ (refer Fig.~\ref{fig:local_layer_planner}(\textbf{C})). If a queried target node $v^{trm}_P$ is present at $v^{tran}_L$, the planner subsequently transitions back to the \textit{Pose} layer, updating the tangible graph representation to $\mathbb{G}_P$ for further planning (refer Fig.~\ref{fig:local_layer_planner}(\textbf{D})). This attribute of the planner, to expand and expose specific local graph representation to meet layer traversal requirements makes it a flexible solution to leverage 3DLSG for path-planning.  

In order to ensure realistic transition between nodes in the highest abstraction layer, we ground the concept of transitioning between $v_T$ nodes within the \textit{Level} and \textit{Pose} layer graphs respectively. Since this work makes use of 3DLSG constructed by an inspect-explore planning autonomy, we syncretize the designation of transition nodes within corresponding local layers with respect to the planning behaviour exhibited in~\cite{viswanathan2024xflie}. Specifically, in order to consider $\pi^G_i$ as completed, the robot has to reach and be localized at $v_{T,i+1}$ within 3DLSG. However, due to the abstract nature of $v_{T,i+1}$ which signifies the occupied position of a target, it is not directly reachable. Therefore, we consider $\pi^G_i$ to be tracked when the robot is localized within the local layers of $v_{T,i+1}$. We designate certain node in \textit{Level} layer, i.e. $v_L(0)$ corresponding to \textit{Level-0} of inspection in this case since subsequent exploration policy is executed at this level~\cite{viswanathan2024xflie}. For the purpose of this work, we consider that the $v_L$ nodes to transition between $v_T$ is designated during 3DLSG construction and is available during path-planning. Within $\mathbb{G}_P$ of $v_L(0)$, we evaluate the transition $v_P$ based on the nearest $v_P$ to $v_{T,i+1}$. This is achieved through a kdtree search over the registered pose attributes of $n_v(\{v_P\})$ with respect to the registered pose attribute of $n_v(v_{T,i+1})$ (refer Fig.~\ref{fig:local_layer_planner}(\textbf{E})). Once the robot navigates towards and is localized at $v_L(0)$ and the evaluated $v_P$ to transition to the next $v_{T,i+1}$, we evaluate the closest $v_P \in \mathbb{G}_P$ of $v_L(0)$ in $v_{T,i+1}$ that can be reached from $v^{curr}_P$. This local route $\pi^L_P$, comprising of the segment from $v^{curr}_P$ in $v^{curr}_L$ of $v^{curr}_T$ to the transition $v_P$ in $v_L(0)$ of $v_{T,i+1}$, is shared with the modified STAGE module for execution. Once the robot is localized within the corresponding layers of $v_{T,i+1}$, $\pi^{G}_i$ is assumed to be completed and the next $\pi^{G}_{i+1}$ is evaluated and shared with the local layer planner (refer Fig.~\ref{fig:local_layer_planner}(\textbf{F})).

The above process is recursively evaluated for each segment of the global layer route until the robot is localized at the queried $v^{trm}_T$ node. Once $v^{curr}_T = v^{trm}_T$, subsequent $v^{tran}_L$ and $v^{tran}_P$ are evaluated based on the queried $v^{trm}_L$ and $v^{trm}_P$.

\begin{figure}[htbp]
    \centering
    \includegraphics[width=\linewidth]{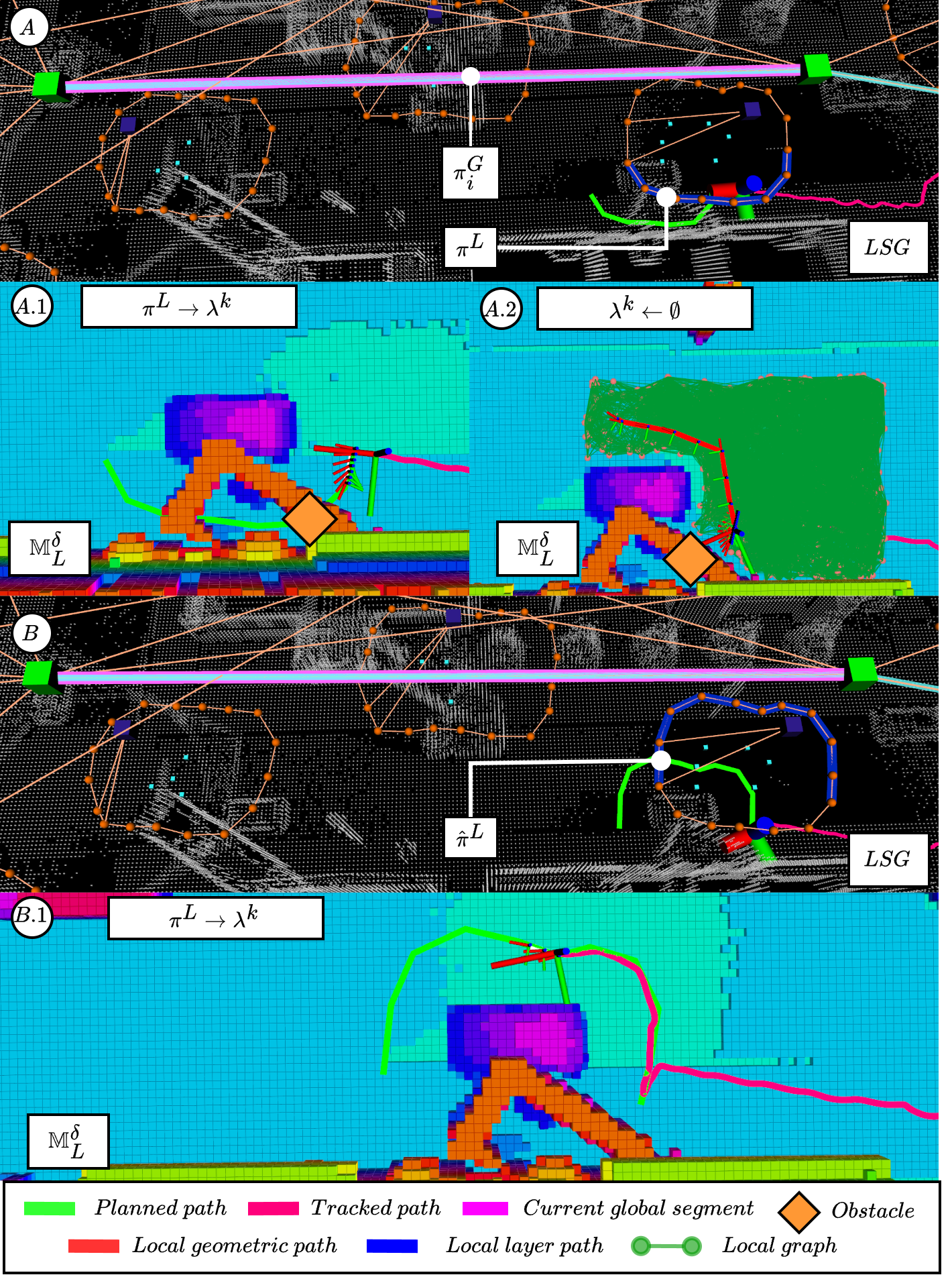}
    \caption{A collage presenting the context of replanning under uncertainty handled by the proposed semantic-geometric path planner.}
    \label{fig:replanning_under_uncertainty}
\end{figure}

\textbf{Replanning under scene uncertainty:} As motivated before, navigating real-world environments necessitates the consideration of uncertainty towards traversability condition of dynamically varying scenes. We introduce ($\Gamma_u \in \mathbb{R}^3$) as the uncertainty in the traversability condition such as path obstructions. This modifies the condition of local map to $\mathbb{M}_L \rightarrow \mathbb{M}^{\delta}_L$. As depicted in algorithm.~\ref{alg:SCOUT} for a given high-level path segment $\pi^{L}$, STAGE checks for free-space traversability in line with the commanded route. In case of detection of obstructions, i.e. occupied voxels lying enroute of $\pi^{L}$, STAGE refines the path segment $\pi^L \rightarrow \lambda^k | \lambda^k \in \mathbb{R}^3$ around the obstacle. In case no feasible alternate solutions exist for STAGE to plan around the obstacle within $\mathbb{M}_L$, the blocked segment is then reported back to the local layer planner. Upon receiving an updated traversability status $\mathcal{T}$ of the respective segment within $\pi^{L}$, SPADE updates the encoded traversability edge attribute within the current $\mathbb{G}^{S}_l$ to reflect the existing situation. Subsequently, the local layer planner shares a replanned $\Hat{\pi}^\mathcal{L}$ with STAGE. In the event no feasible solutions during replanning within $\mathbb{G}^{S}_l$, i.e $\pi^{L} \gets \emptyset$, the encoded $\mathcal{T}$ of the edge within $\mathbb{G}_T$ connecting the corresponding current global route segment $\pi^G_i$ is updated to reflect the presence path observations. At this point, a replanning on the global layer occurs to provide an alternate global layer route to the queried nodes ($\{\Hat{\Pi^G}\}$). Figure.~\ref{fig:replanning_under_uncertainty} presents a visual collage of the replanning process when encountering path obstructions due to dynamic scene conditions in a simulation environment. In Fig.\ref{fig:replanning_under_uncertainty}(A), the current global route segment $\pi^G_i$ to be traversed is highlighted in response to a navigational query. The corresponding local layer route $\pi^L$ is planned to $v^{tran}_P$ from $v^{curr}_P$ in order to transition to $v_{T,i+1} \in \pi^G_i$. Figure.~\ref{fig:replanning_under_uncertainty}(A.1) captures the change in the volumetric map reflecting the introduced path obstructions along the way of the initial $\pi^L$. As the robot nears the obstacle and a probable collision is perceived through collision checks along the way, the $\pi^L$ is discarded and the alternate local routes are evaluated in Fig.~\ref{fig:replanning_under_uncertainty}(A.2). Figure.~\ref{fig:replanning_under_uncertainty}(B) shows $\Hat{\pi}^L$, the replanned high-level route once the traversability status of the leading edge lying in untraversable space is updated. This is then subsequently shared with the geometric planner to navigate the scene in Figure.~\ref{fig:replanning_under_uncertainty}(B.1). Thus, replanning an alternate route and going around the obstruction.

\section{Setup and evaluation}

We evaluate SPADE architecture in both large-scale outdoor urban simulation environment as well as experimentally in a outdoor urban environment deployment for autonomous inspection mission. We mainly evaluate the architecture on two main fronts: (a) To address run-time navigation queries through the integrated xFLIE planner, i.e. navigating towards the next inspection target in addition to addressing user-defined navigational queries. (b) To adapt to traversability uncertainty such as path obstructions in dynamic scenes. The architecture is evaluated onboard Boston Dynamics (BD) Spot quadruped robot equipped with Realsense D455 stereo-camera, Vectornav VN-100 Inertial Measurement Unit (IMU) and Velodyne High-res VLP16 3D LiDAR. We use Nvidia Jetson Orin embedded computational board running ROS Noetic, Ubuntu 20.04 and JetPack 5.1. For simulation, we use the RotorS~\cite{Furrer2016} simulator along with Gazebo and ROS Noetic on Ubuntu 20.04 LTS operating system with i9-13900K CPU and 128 GB RAM. Additional results on the evaluation of SPADE can be accessed through \url{https://sites.google.com/view/spade-planner/home}.

\section{Results and Discussions}

\begin{figure}[hpbt]
    \centering
    \includegraphics[width=0.8\linewidth]{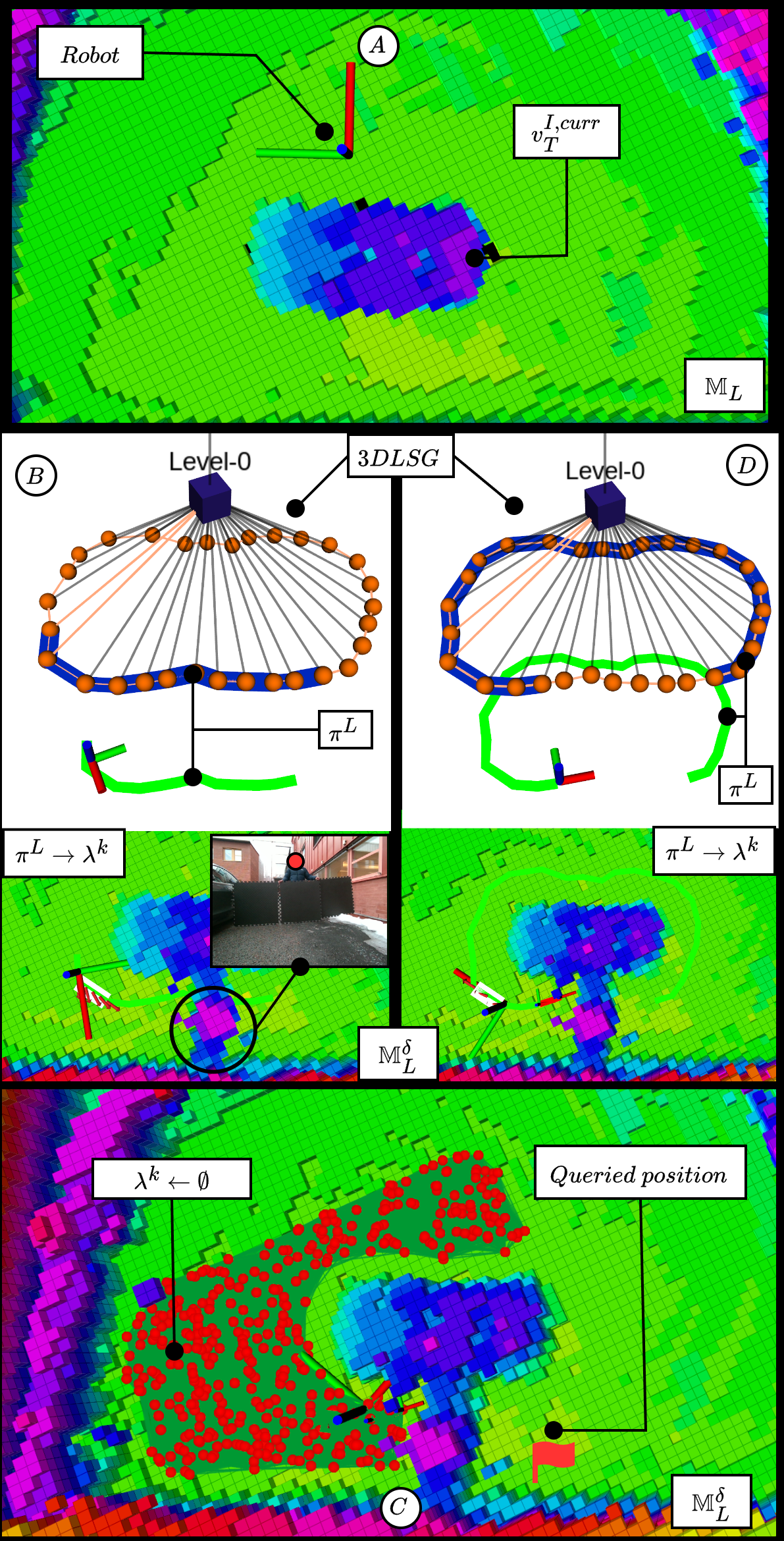}
    \caption{A collage capturing the adaptive response of the semantic-geometric planner under introduced uncertainty to the operating environment.}
    \label{fig:exp_dyn_replan}
\end{figure}

Figure.~\ref{fig:exp_dyn_replan} presents the evaluation of planning under uncertainty during the experimental run. Fig.~\ref{fig:exp_dyn_replan}(A) presents the initial $\mathbb{M}_L$ built during autonomous inspection. At the end of inspection, an operator-defined navigation query is sent to SPADE. Fig.~\ref{fig:exp_dyn_replan}(B)(\textit{On top}) presents the local layer route planned across $\mathbb{G}_P$. (\textit{On bottom}), the top-down view of the volumetric map is shown with the introduced uncertainty highlighted ($\mathbb{M^{\delta}_L}$). As $\pi^L$ is processed and tracked by the local geometric planner, the traversable status of the initial route is changed once the path obstruction is perceived through collision-check. Subsequently, Fig.~\ref{fig:exp_dyn_replan}(C) highlights the evaluation of an alternate traversable route (in \textit{green}) by the geometric planner to reach the queried position. Once no alternate solutions are found, the local layer planner updates the traversable status of the immediate edge that is evaluated to be currently untraversable. Finally, Fig.~\ref{fig:exp_dyn_replan}(D) present the re-planned route across $\mathbb{G}_P$ to reach the queried pose. This is then shared with the geometric planner for tracking. 

\begin{figure*}[!ht]
    \includegraphics[width=\linewidth]{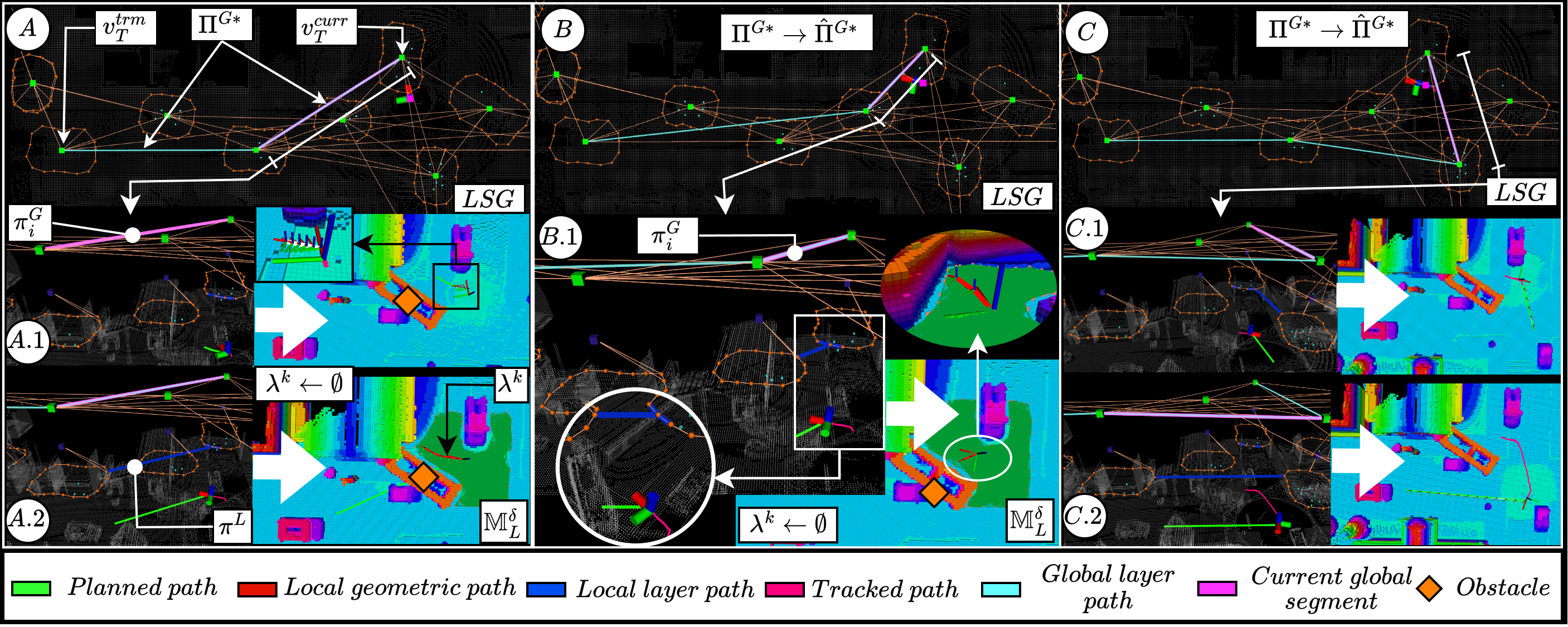}
    \caption{A visual collage capturing large-scale simulation where global layer replanning is highlighted.}
    \label{fig:sims_global_planning}
\end{figure*}

Figure.~\ref{fig:sims_global_planning} presents the evaluation of SPADE within the urban city simulation environment. For the mission, the robot was tasked to inspect vehicles. During the mission, xFLIE planner raises navigational query to get to the next object of inspection. In Fig.~\ref{fig:sims_global_planning}(A), the path-planner is queried to visit $v^{trm}_T$. As discussed above, SPADE localizes the robot within the abstraction layers of LSG and subsequently generates a global and the corresponding local layer route (refer Fig.~\ref{fig:sims_global_planning}(A.1-A.2). However, as indicated by the inset \textit{top-down} view of the volumetric map, the planned route encounters path obstructions. Once there are no feasible alternate routes by both local geometric planner and within the local layers ($\pi^L,\lambda^k \gets \emptyset$), the $\pi^G_i$ is discarded and a global layer replanning occurs (refer Fig.~\ref{fig:sims_global_planning}(B)). Figure.~\ref{fig:sims_global_planning}(B.1) highlights the updated $\Hat{\pi}^G_i$ and subsequent evaluation of the local layer routes. However, as indicated, the obstructed local scene leads to the current local routes being discarded and $\mathbb{G}_T$ updated on the traversability status. This leads to a third replanning instance on the global layer as seen in Fig.~\ref{fig:sims_global_planning}(C). In Fig.~\ref{fig:sims_global_planning}(C.1), the planner finds a feasible alternate global route going around the obstacle and plans subsequent local layer routes to traverse the current global route segment. As the robot tracks the path, Fig.~\ref{fig:sims_global_planning}(C.2) showcases the evaluation of the next global route segment and the corresponding local layer route to navigate through the scene. 

Thus, we show that through synchronous local planning, in the context of abstract and local geometric scene representation, SPADE attempts to exhaust potentially valid local layer routes before replanning on the global layer and trying again. This behaviour underscores the ability of SPADE to plan and effectively adapt to dynamic changes in the environment without needing to reinitiate single-step global replanning at every instance of encountering path obstructions.

\textbf{Performance Comparison:}~
\begin{figure}[htbp]
    \centering
    \includegraphics[width=0.98\linewidth]{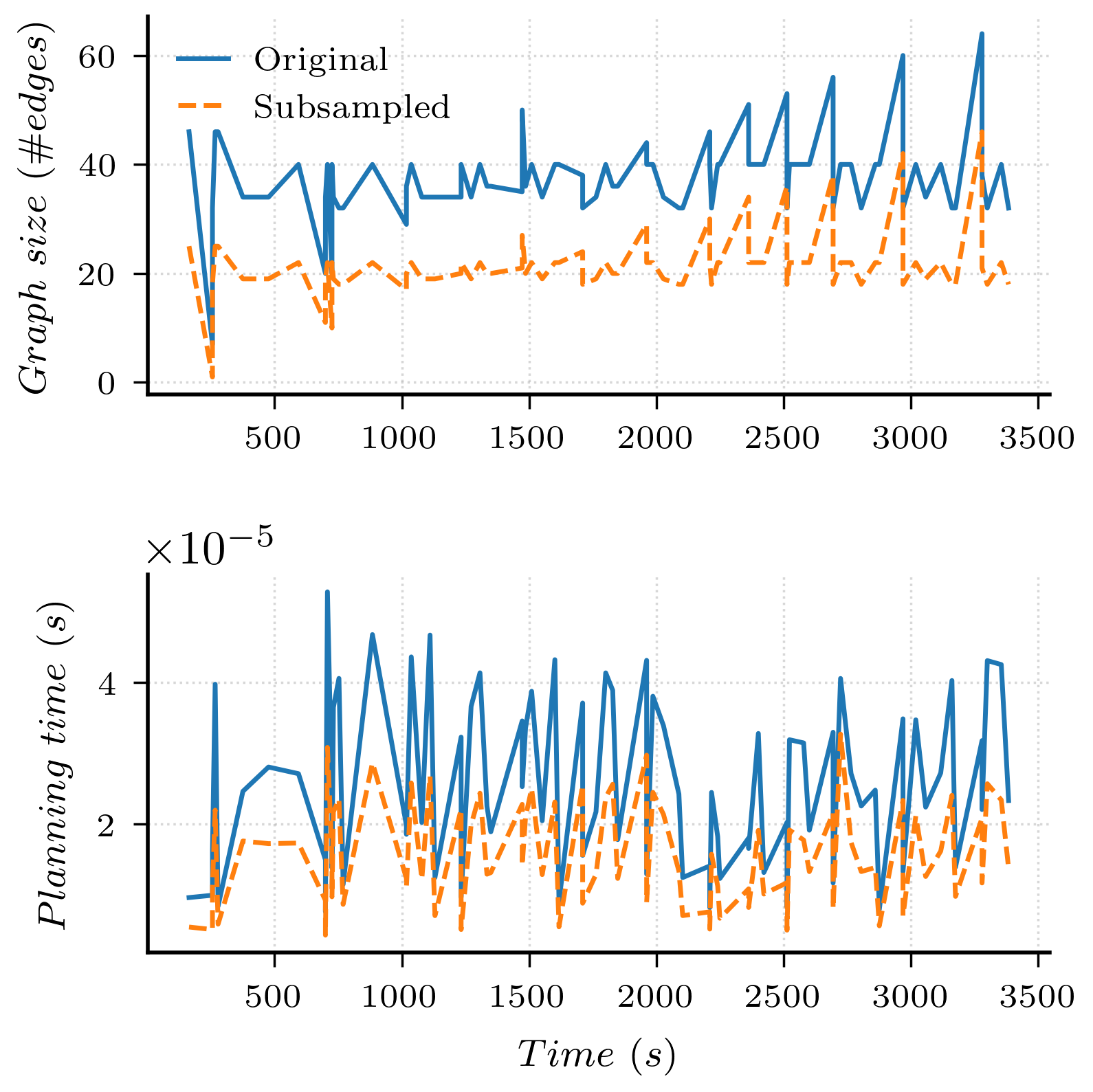}
    \caption{Planning metrics capturing the evolution of (\textit{On top}) the graph size processed by the graph-based path planner and (\textit{On bottom}) the corresponding path computation time.}
    \label{fig:spade_metrics_sims}
\end{figure}
Figure.~\ref{fig:spade_metrics_sims} presents a quantitative analysis of SPADE performance in computing a traversable route within a domain subsampled graph ($\mathbb{G}^S_l$) against an original graph topology ($\mathbb{G}_l$). As motivated previously, we consider multiple task-specific edges capturing various relational information from the environment, present within the 3DLSG. Specificially, as seen in Fig.~\ref{fig:spade_metrics_sims}(\textit{top}), the graph size, i.e. the number of edge within a graph, is higher than that of the subsampled graph when processed by SPADE for the path-planning task. The effect of this can be visualized within the performance plot of the computation time required to plan a path in Fig.~\ref{fig:spade_metrics_sims}(\textit{bottom}). Although the recorded computation time of both methods is on the order of milliseconds, the relative improvements in planning time achieved using a subsampled graph demonstrates the benefits of leveraging domain-specific graphs for specific tasks. The subsequent peaks at later time-steps correspond to the navigation query requiring the robot to navigate through the large urban city to continue inspection. The rising peaks near the end reflect the increase in the relative traversable edges established between inspected targets as the robot covers and maps more of the environment. 

\begin{figure}
    \centering
    \includegraphics[width=\linewidth]{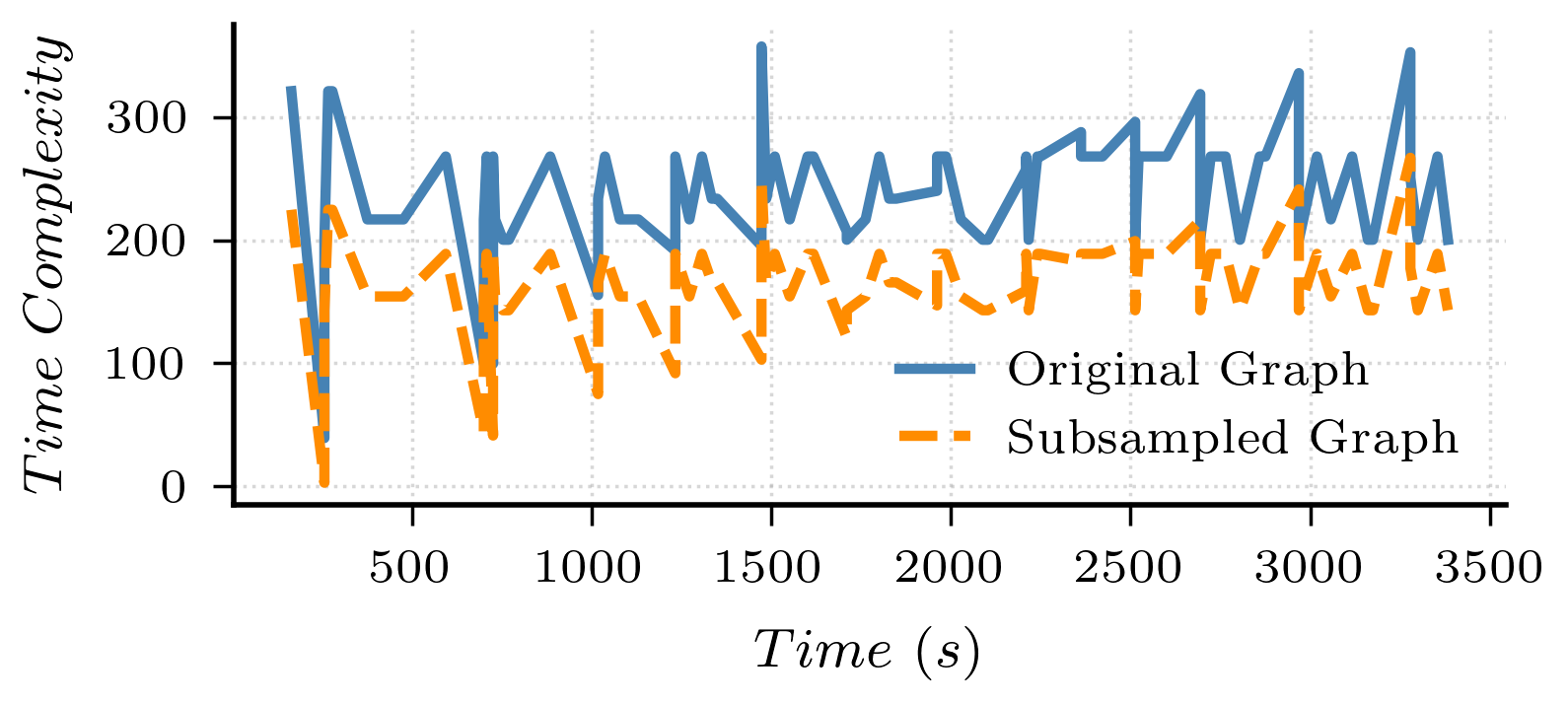}
    \caption{A time complexity analysis between planning on a domain-sampled graph and an original graph.}
    \label{fig:time_complex}
\end{figure}

Figure.~\ref{fig:time_complex} highlights the computation efficiency achieved by processing a subsampled graph during path planning compared to planning over a naive original graph. Since we used a priority queue implementation to model Dijsktra's graph based path planner, we plot the overall time complexity model $\mathcal{O}((|E|+|V|)log_2|V|)$ in Fig.~\ref{fig:time_complex}. Currently, the original graph available are composed of two edge types combinations: (a) \textit{observational/traversable} within $\mathbb{G}_T$ and (b) \textit{symbolic/traversable} within $\mathbb{G}_P$. Based on the data, we validate that the domain sampling improves the computational efficiency to process a graph for extracting a traversable route.  If the internal composition of these layer representations evolve to consider other edge types, the theoretical time complexity increases when planning over an non-subsampled graph and by extension the required computational time. In contrast, the implementation of planning with domain-subsampling would remain agnostic to the evolution and could ensure scalability to execute a required task over a graph when expanding to larger scenes.  
 
\section{Conclusions}

This work presents the SPADE architecture: a tightly coupled semantic-geometric path planner designed to leverage the ease of initial planning over abstract representations while integrating occupancy based local planner grounded in the volumetric representation of the environment for collision-free navigation. By bifurcating the planning problem depending on the constituent layers, SPADE prioritizes local scene navigation, in the aspect of abstract representation as well as geometric scene model. Furthermore, we demonstrate the impact subsampling multi-domain scene graphs can have on path planning performance with a demonstrated $\approx 40 \%$ faster computation of path by a graph-based planner compared to planning on a non-subsampled graph. Additionally, we highlight the adaptability of SPADE to address navigation in dynamic environment conditions that could introduce uncertainty to the traversability condition on previously built 3DLSG. As part of future works, we highlight the integration of SPADE with other defined tasks such as scene reasoning and object search, underscoring the impact of domain-aware subsampling of graphs. 
\bibliography{bib}

\end{document}